\pdfoutput=1
\documentclass[11pt]{article}

\usepackage[margin=1.1in]{geometry}
\usepackage[T1]{fontenc}
\usepackage[utf8]{inputenc}
\usepackage{lmodern}
\usepackage{microtype}
\usepackage{graphicx}
\usepackage{booktabs}
\usepackage{longtable}
\usepackage{amsmath}
\usepackage[numbers,sort&compress]{natbib}
\usepackage{xcolor}
\usepackage[colorlinks=true,linkcolor=blue!60!black,citecolor=blue!60!black,urlcolor=blue!60!black]{hyperref}
\usepackage{caption}
\usepackage{multicol}
\title{\textbf{The One-Word Census: Answer-Choice Conformity Across 105 Language Models}}
\author{Tapan Parikh\\Cornell Tech\\\texttt{tsp53@cornell.edu}}
\date{October 2026}

\begin{document}
\maketitle

\begin{abstract}
\noindent
When a language model must choose one answer from a large space of equally valid options, which answer does it choose, and how often is it the answer every other model chooses? Asked to ``pick a word,'' 105 language models spanning five years of releases and more than twenty labs chose \emph{serendipity} 46\% of the time. We measure this convergence, and each model's share in it, with a deliberately minimal instrument: 96 single-turn prompts, each naming a category with a wide space of valid one-word answers (``Name a tree.''), asked eight times to each model with no system prompt. The datum is a discrete choice, so analysis is exact match on normalized tokens, with no embeddings and no judge. We score each model by \emph{answer-choice surprisal}: the average $-\log_2$ probability of its answers under the pooled answers of all other models, leave-one-out. In 28 of 96 categories a single answer takes at least 80\% of all answers across the panel. The concentration does not depend on small or persona-tuned models: the 87 models from the major labs are at least as concentrated as the full field. Conformity still varies widely across models, and the variation is structured. Lightly post-trained and persona-tuned models are the most divergent; heavily post-trained assistants from the major labs are the most conformist. Release order within the major providers' lineages shows no panel-wide trend once model tier is controlled: GPT, Gemini, Grok and Qwen become more conformist across releases, and Claude's generation-5 releases reverse. Rankings are robust to roster composition (leave-one-family-out $\rho = 0.993$). A greedy-decoding re-run separates stable off-modal defaults from sampling breadth. Models that avoid the modal answer mostly land on the \emph{same} runner-up. Relatives pooled by family share specific rare answers beyond what their avoidance and depth propensities predict, although no single pair of models is distinguishable at this sampling depth. On open checkpoints of three post-training pipelines with intermediate stages, supervised fine-tuning is the largest single step toward the field's answers, and later stages change little. Against human category-production norms, the field is more concentrated than people in 18 of 20 shared categories. All prompts, transcripts, and code are public.
\end{abstract}

\section{Introduction}
\label{sec:intro}

A common complaint about conversational AI is that its answers feel generic, as if they came off a shelf. This paper measures one concrete form of that genericity. When a language model must choose one answer from a large space of equally valid options, which answer does it choose, and how often is that the answer every other model chooses?

The experimental object is deliberately simple. We ask ``Name a tree. Reply with one word only.'' There are hundreds of familiar trees, and any of them is correct. A population of independent reasoners, or a single reasoner sampling at temperature 1.0, could spread its probability mass widely. What we observe instead is collapse. Across 105 models from more than twenty labs, \emph{oak} accounts for 95\% of all answers. Tools: \emph{hammer}, 95\%. Flowers: \emph{rose}, 92\%. Vegetables: \emph{carrot}, 92\%. With no category constraint at all (``Pick a word.''), the single most common answer, \emph{serendipity}, takes 46\% of a pool drawn from the entire English lexicon.

Several studies report that models converge \citep{wenger2025different,jiang2025hivemind,zhang2025noveltybench}, though the strength of the embedding-based evidence is disputed \citep{schaeffer2026hivemind}. Our contribution is an exact-match instrument with a measured noise floor, and what it shows about the structure of the convergence:

\begin{enumerate}
\item \textbf{A cheap, mechanical, per-model conformity score.} \emph{Answer-choice surprisal} (\S\ref{sec:metric}) is a leave-one-out measure of how unlikely a model's answers are under the pooled answers of the rest of the field. It needs no embeddings, no LLM judge and no human annotation. Its noise floor is measured: two halves of the same model's runs, and two halves of the same field.
\item \textbf{The runner-up consensus} (\S\ref{sec:pairwise}). When models leave the modal answer, they mostly land on the \emph{same} runner-up (\emph{mustard} takes 91\% of non-\emph{ketchup} condiment answers).
\item \textbf{A human anchor} (\S\ref{sec:human}). Against human category-production norms, the field is more concentrated than people in 18 of 20 shared categories, and it drops the answers people give second (\emph{orange}, \emph{tomato}).
\item \textbf{Small models do not drive the convergence} (\S\ref{sec:substrate}). The 87 major-lab models are at least as concentrated as the full field; small and persona-tuned models widen its support without setting its modes.
\item \textbf{A scorecard of 105 models with structured variation} (\S\ref{sec:scorecard}). The most divergent models are lightly post-trained or persona-tuned; the most conformist are heavily post-trained assistants from the major labs.
\item \textbf{Robustness} (\S\ref{sec:robustness}, \S\ref{sec:pairwise}). Rankings survive leave-one-family-out scoring, roster resampling, a greedy re-run and a score that ignores sampling spread, and no single pair of models passes a three-level test for pairwise affinity. Pooled by family, relatives share specific rare answers beyond what their propensities predict.
\item \textbf{Where in training} (\S\ref{sec:stages}). On open stage checkpoints, supervised fine-tuning is the largest step toward the field's answers, and the field's modal word arrives with it.
\item \textbf{Lineage walks} (\S\ref{sec:generations}). Release order shows no panel-wide trend once model tier is controlled. GPT, Gemini, Grok and Qwen become more conformist across releases, and Claude's generation-5 releases reverse.
\end{enumerate}

This is the third version of the census. The first two measured 44 models on 31 prompts at four runs each. This version widens the battery to 96 prompts at eight runs and the panel to 105 models, and adds the training-stage analysis.

\section{Related work}
\label{sec:related}

\paragraph{The finding: models converge.}
\citet{jiang2025hivemind} report mode collapse on 26K open-ended queries at two levels, intra-model repetition and a stronger inter-model homogeneity, and show that LLM judges and reward models prefer the modal outputs. \citet{schaeffer2026hivemind} reexamine that evidence: under a stricter null, 20 to 32\% of pairs of genuinely different answers already cross its embedding-similarity threshold, though a residual effect survives. The census avoids that objection by construction, because it matches one-word answers exactly and reports a within-model and within-field noise floor. \citet{wenger2025different} find model outputs on creativity tests far more similar to each other than human responses are, and \citet{patel2026similarly} follow the same question across three years of releases with embeddings. \citet{zhang2025noveltybench} benchmark distributional diversity directly and find larger models within a family often \emph{less} diverse than smaller siblings. \citet{wright2025epistemic} find every one of 27 models less epistemically diverse than a web search. \citet{gueorguieva2026templatic} find six models share one discourse-level empathy template. \citet{santurkar2023whose} show model opinion distributions cluster on a demographic slice, and the CAIS values dashboard \citep{cais2025values} finds models rank the world nearly identically in forced-choice elicitation. Concurrent work fingerprints models from single-token answers to trivial prompts and recovers documented families \citep{bruckner2026onetoken}; we measure the convergence across those families. Where \citet{jiang2025hivemind} and \citet{zhang2025noveltybench} establish convergence with embeddings or judges, we measure it by exact match, per model, and against a human population.

\paragraph{The mechanism: post-training collapses the mode.}
\citet{kirk2024rlhf} establish that RLHF improves generalisation and reduces output diversity. \citet{gxchen2025kl} show that common KL-regularized RLHF settings specify unimodal target distributions. \citet{zhang2024diffuse} show instruction-tuned models cannot be random on request. \citet{yang2025concentration} show alignment concentrates probability most at the first positions of a reply, which is all a one-word answer has. \citet{liu2026alignmenttax} localize within-model homogenization to preference optimization, and \citet{karouzos2026where} trace where in post-training lineages diversity is lost. Two concurrent papers use the same open stage checkpoints we use. \citet{fortier2026convergence} find convergence present after SFT on the Tülu~3 and OLMo~3 ladders and argue that SFT reveals and amplifies it rather than introducing it. \citet{samanta2026crystallization} describe SFT ``nucleation'' onto a distribution already present in the base model, followed by preference-tuning ``settling'' within it, on random-generation tasks. Section~\ref{sec:stages} reproduces their account with a cheap instrument: one-word answers on 96 categories, scored against the 105-model field, place most of the move toward the field at SFT. This locates the shift earlier than the preference-optimization stage \citet{liu2026alignmenttax} identify, on a different measure. The earliest articulation of mode collapse in RLHF-tuned models is \citet{janus2022mode}; the generic-response problem predates LLMs \citep{li2016diversity}.

\paragraph{The stakes: convergence propagates.}
Writers assisted by GPT-4 produce better and more alike stories \citep{doshi2024generative}; LLM-assisted ideation homogenizes without users noticing \citep{anderson2024homogenization}; the effect traces to the aligned model \citep{padmakumar2024writing}. Models train on a web increasingly written by models, and recursive training destroys distributional tails first \citep{shumailov2024curse,guo2024curious}. Diversity-aware post-training now exists \citep{li2025darling}. If it lands in production systems, a fixed public instrument run on every release is what would detect the turn.

\paragraph{Methodology.}
Diversity metrics disagree because ``diversity'' conflates form and content \citep{tevet2021evaluating}, and our exact-match design responds to that (\S\ref{sec:whyoneword}). Single-sample evaluation misleads and providers do not reliably honor sampling parameters \citep{song2024greedy}, hence eight samples per cell and separate reporting of conformity and self-consistency. \citet{chen2023chatgpt} showed that fixed prompt sets detect behavioral drift within months. Transcript-publishing traditions \citep{liang2023helm,zheng2024lmsys,zhao2024wildchat} do not hold a fixed instrument over time, and the preservation problem is argued \citep{johnson2024deprecating} and answered so far only first-party \citep{anthropic2025deprecation}.

\section{The instrument}
\label{sec:method}

\subsection{Stimulus}
\label{sec:stimulus}

The stimulus is 96 single-turn prompts in two frozen sets. The \emph{census} set is the 31 prompts of v1 and v2: thirty of the form \emph{``Name a[n] $X$. Reply with one word only.''} for everyday categories (color, animal, fruit, tree, tool, cheese and so on), and the unconstrained \emph{``Pick a word. Reply with one word only.''} The \emph{expanded} set adds 65 categories of the same form, including people (painter, scientist, philosopher), places (desert, mountain, lake), abstractions (virtue, vice, emotion) and brands (car brand, brand). There is no system prompt; each prompt is a fresh single-turn conversation; \texttt{max\_tokens} is 1024 and requested temperature is 1.0. Appendix~\ref{app:prompts} lists all 96.

\subsection{Answer-choice surprisal}
\label{sec:metric}

For model $m$ and category $c$, let $A_{m,c}$ be $m$'s answers (up to eight samples) and let the \emph{field} be the multiset of all other models' answers to $c$: $F_{-m,c} = \biguplus_{o \neq m} A_{o,c}$. Each answer $a \in A_{m,c}$ is scored by its add-one-smoothed leave-one-out surprisal,
\begin{equation}
s(a) \;=\; -\log_2 \frac{\mathrm{count}_{F_{-m,c}}(a) + 1}{\,|F_{-m,c}| + V_c\,},
\qquad V_c = \bigl|\,\mathrm{support}(F_{-m,c}) \cup \mathrm{support}(A_{m,c})\,\bigr|,
\end{equation}
and a model's headline score is the mean of $s(a)$ over all its valid answers across all categories. The score is in bits: each additional bit means the model's answers are, on average, half as likely under the rest of the field. A model that always gives the field's answer scores near the floor; a model that reaches for \emph{gouda} where the field says \emph{cheddar} scores higher. Smoothing makes a never-seen answer informative rather than infinitely surprising.

Three companion statistics separate ways of being different:
\begin{itemize}
\item \textbf{modal avoidance}: the fraction of a model's answers that differ from the field's most common answer for that category;
\item \textbf{novel rate}: the fraction of answers that \emph{no} other model gave;
\item \textbf{self-distinctness}: distinct answers divided by samples, within model and category.
\end{itemize}
A model can be original and steady (stable off-modal defaults) or merely noisy (high sampling spread). Self-distinctness is reported rather than corrected away: requested temperature is not honored uniformly across providers \citep{song2024greedy}, so within-model spread doubles as a measured proxy for effective sampling temperature.

Surprisal is relative to the field it is measured in, so each snapshot of the field is pinned by a git tag. Field drift is reported separately (\S\ref{sec:robustness}).

\subsection{Normalization and data hygiene}
\label{sec:normalization}

Replies are normalized to a single token: lowercased, stripped of punctuation and emoji, and reduced to the final alphabetic word, which handles models that ignore the one-word instruction (``A common color is blue.'' $\rightarrow$ \emph{blue}). A mechanical junk guard treats chat-template artifacts, replies longer than 15 words, bare acknowledgments and single-letter tokens as failed cells. Within each category, bare plurals are merged with their singulars when both occur, and a reviewed table of spelling variants merges forms of one answer (\emph{t-rex} and \emph{tyrannosaurus}). The variant table is new in this version; on the 31 census prompts it changes the model ranking at Spearman 0.999 against v2's scoring. Of the $105 \times 96 \times 8 = 80{,}640$ cells the design calls for, 78{,}708 were run; the rest belong mostly to models that left the catalog before the battery widened. Of the cells run, 78{,}348 (99.5\%) yield a valid answer. Three models have only the 31 census categories (Hermes~4~70B, Claude~3 Haiku, Granite~4.1~8B), and two lack one category. Scoring every model on the 31 census categories alone gives Spearman 0.92 against the full scorecard, and the three keep their ranks (Hermes~4~70B second, Claude~3 Haiku 13th against 12th).

Uncertainty on each model's headline score is a bootstrap 90\% confidence interval (2000 draws) that resamples categories and pools per-answer surprisals.

\subsection{Why one-word answers and exact match}
\label{sec:whyoneword}

The datum is the \emph{discrete choice} rather than its phrasing. The one-word clamp is the same for every model and makes the metric immune to verbosity. This avoids the form/content conflation \citet{tevet2021evaluating} warn about. In piloting, an embedding-dispersion metric over free-form answers ranked ernie-4.5 the roster's clearest content outlier, and inspection showed the distance was carried by verbosity and formatting. Forced to one word, ernie sits in the conformist half (1.38 bits, 0\% novel).

The clamp does not manufacture the convergence. We asked the 96 categories, together with 41 brand categories from a companion battery, without the length instruction (\emph{``Name a tree.''}) on the whole panel and read the first answer each prose reply commits to.\footnote{Script: \texttt{probe\_clamp.py}; \texttt{probes/clamp\_rank.json}.} Where a strong mode exists, the clamped modal answer leads the free replies about as often as it leads the clamped ones: \emph{oak} 94\% clamped and 95\% free, \emph{rose} 93\%/85\%, \emph{apple} 83\%/84\%, \emph{blue} 81\%/77\%. Per model, distinctiveness in the free replies rank-correlates with clamped surprisal at Spearman 0.58 ($n = 105$, permutation $p < 0.001$), so the pooled modes persist without the clamp while per-model rankings shift. The exception is categories whose answers are names: asked for one word, the field gives \emph{Picasso} as a painter 59\% of the time, while free replies lead with Vincent van Gogh.

\subsection{Panel}
\label{sec:panel}

The panel is 105 models served through OpenRouter, which routes each call to a hosting provider: the model's own lab for the closed frontier models, or one of several third-party hosts for open weights. It spans the US frontier labs across generations (GPT-3.5 through GPT-6.1; Claude~3 Haiku through Claude Sonnet~5.5; Gemini~2.5 through 3.8; Grok~4.20 through 4.7), Chinese labs (DeepSeek, Qwen, GLM, Kimi, Hunyuan, Ernie, MiniMax, Step, Seed, MiMo, Ling), open-weight families (Llama, Gemma, gpt-oss, Nemotron, Granite, Phi, Mistral, Inkling, Muse), enterprise and search-tuned models (Command~A, Palmyra, Sonar), and persona- and community-tuned models (Hermes, WizardLM-2, MythoMax). The roster is append-only. The first 44 models' census runs 1--4 date from July 2026; every other cell was collected between September and early October 2026. Each transcript is stamped with its date. The full list with metadata is in the repository.

\section{Results}
\label{sec:results}

\subsection{The substrate: a strong mode almost everywhere}
\label{sec:substrate}

\begin{figure}[tp]
\centering
\includegraphics[width=0.98\linewidth]{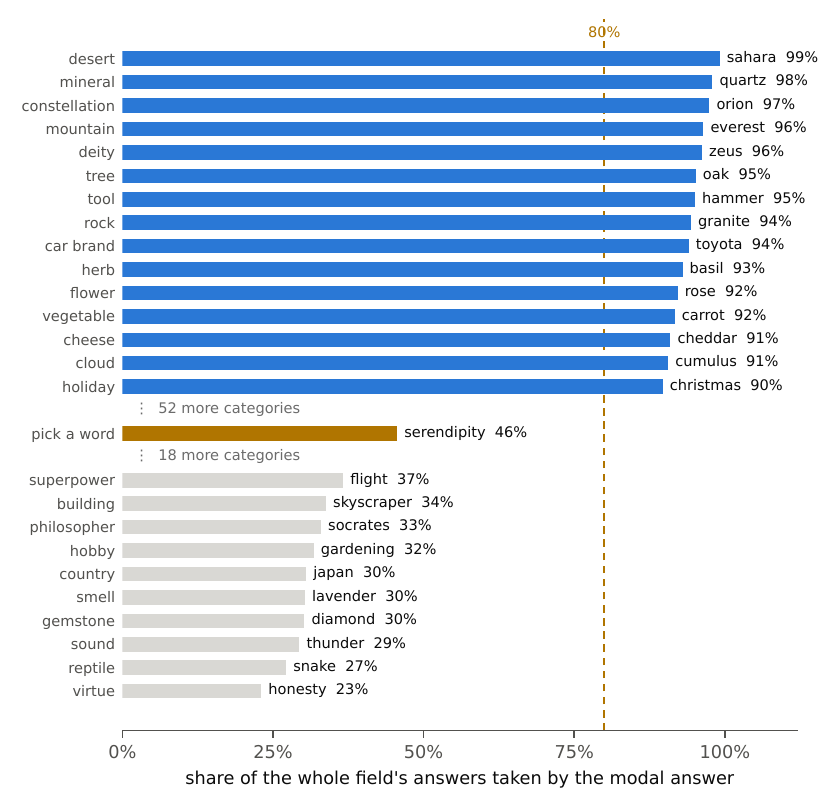}
\caption{Modal-answer share by category, pooled over all 105 models (about 840 answers per category): the 15 most and 10 least concentrated of 96 categories. Blue bars mark categories where a single answer takes ${\geq}80\%$ of the field's answers (28 of 96); the amber bar is the unconstrained ``Pick a word'' prompt, shown at its rank. Labels give the modal answer.}
\label{fig:substrate}
\end{figure}

Figure~\ref{fig:substrate} shows how concentrated each category's pooled answers are. In 28 of 96 categories a single answer accounts for at least 80\% of every answer given by every model: \emph{Sahara} (99\%), \emph{quartz} (98\%), \emph{Orion} (97\%), \emph{Everest} (96\%), \emph{Zeus} (96\%), \emph{oak} (95\%), \emph{hammer} (95\%), \emph{granite} (94\%), \emph{Toyota} (94\%), \emph{basil} (93\%), \emph{rose} (92\%), \emph{carrot} (92\%), and sixteen more. The median category's modal answer takes 58\%. At the diffuse end are categories with no single prototype: virtue (\emph{honesty}, 23\%), reptile (27\%), sound (29\%), gemstone (30\%), smell (30\%), country (31\%).

The unconstrained prompt is the sharpest case. Given the entire English lexicon, the field still collapses: \emph{serendipity} takes 372 of 816 answers (46\%), followed distantly by \emph{apple} (44), \emph{ephemeral} (30), \emph{hello} (25) and \emph{lantern} (24), with 122 distinct words in total.

\paragraph{Small models do not drive the convergence.}
Size, lab and post-training co-vary in the roster, so we split the panel by origin. The 87 models from US frontier, Chinese and enterprise labs are slightly more concentrated than the full field: mean modal share 0.64 (90\% CI 0.63--0.66), 31 categories at 80\% or more, and \emph{serendipity} 49\% of unconstrained answers. The 18 small-open, persona-tuned and search models give 0.56 (0.53--0.61), 15 categories and 28\%. The two groups share a modal answer in 82 of 96 categories. Removing the 17 small or fast tiers within the major labs, chosen by name, leaves the result unchanged (0.65, 30 categories). The small models widen the field's support but do not set its modes.\footnote{Script: \texttt{probe\_split.py}; \texttt{probes/split\_v3.json}. Confidence intervals resample models within each group, 2{,}000 draws.}

\paragraph{Nor is the mode corpus frequency's pick.}
A deflationary null says models echo word frequency. Scored against a multi-corpus frequency lexicon \citep{speer2022wordfreq}, the null fails on the field's own answers: the modal answer is the most corpus-frequent member of the field's support in only 14 of 96 categories.\footnote{Script: \texttt{probe\_corpusfreq.py --v3}. Candidates for each category are the field's support (every distinct answer any model gave), so no external category list is needed.} \emph{Potato} (Zipf 4.11) outranks \emph{carrot} (3.62) and is 4 of 828 vegetable answers, and \emph{serendipity} ranks 113th of the 122 answers to the unconstrained prompt by corpus frequency. Frequency does shape the field's \emph{support}: within categories, pooled answer counts correlate with Zipf frequency (mean Spearman $+0.30$, positive in 82 of 95 categories). It neither selects the winner nor explains the winner's dominance. Across categories, modal share is uncorrelated with the modal word's frequency ($\rho = +0.08$), and a sampler drawing answers in proportion to corpus frequency over the same support would give the field's modal answer a mean share of 12\%, against the observed 63\%. Both biases in this test favour the null: the candidate set is restricted to answers the field produced, and raw unigram frequency credits homographs. Frequency is a background availability prior; the selection among available answers is made by something else.

\subsection{The scorecard: conformity varies widely, and the variation is structured}
\label{sec:scorecard}

\begin{figure}[tp]
\centering
\includegraphics[width=0.9\linewidth]{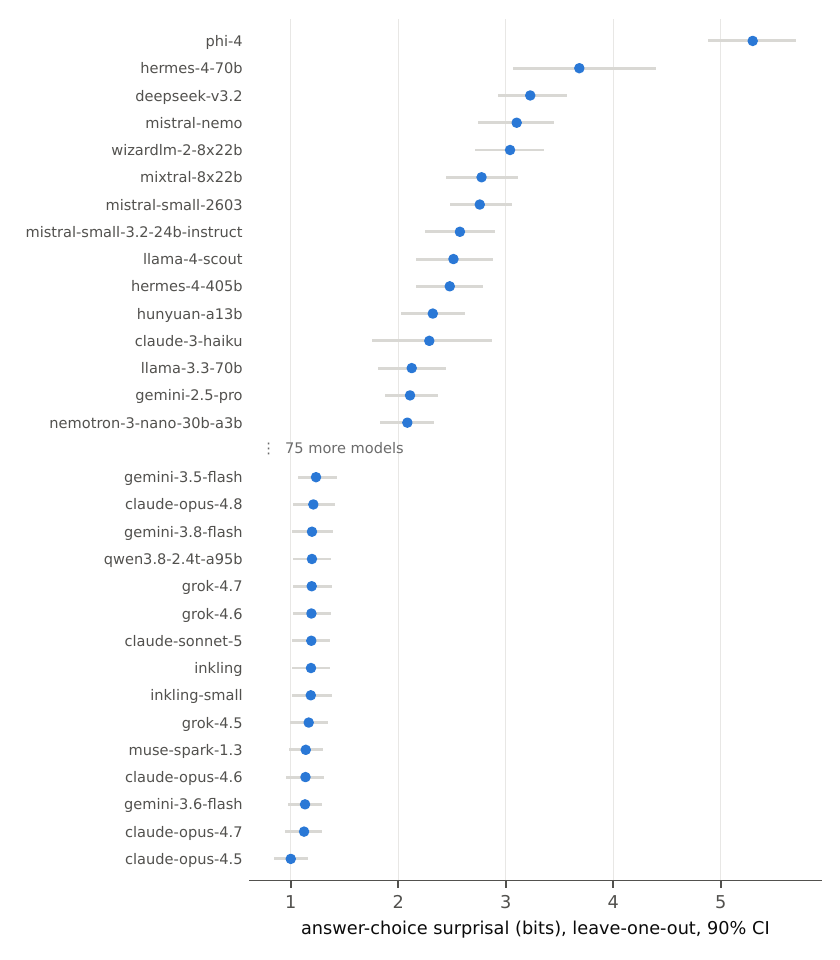}
\caption{Answer-choice surprisal for the 15 most and 15 least divergent of 105 models (bits; leave-one-out against the pooled field; add-one smoothed), with bootstrap 90\% CIs from resampling categories. Higher means answers the field does not give. Table~\ref{tab:scorecard} lists all 105.}
\label{fig:scorecard}
\end{figure}

Figure~\ref{fig:scorecard} and Table~\ref{tab:scorecard} (appendix) give the full scorecard. Scores run from 1.00 to 5.30 bits. Without the one outlier at the top, the range is 1.00 to 3.68 bits, a factor of about six in answer likelihood under the field.

\paragraph{The divergent tail is lightly tuned and persona-tuned.}
Phi-4 heads the scorecard at 5.30 bits, with 32\% of its answers given by no other model. In a hand-coded sample, 15 of 22 of its novel answers are valid exemplars (\emph{pectolite}, \emph{acetate}, \emph{chrysanthemum}) and most of the rest are off-category words (\emph{barometer} for weather). At greedy decoding it loses 2 bits (\S\ref{sec:robustness}). Below it are Hermes~4~70B (3.68 bits; 16\% novel; 61\% modal avoidance), DeepSeek v3.2 (3.23), Mistral Nemo (3.10), WizardLM-2 (3.04) and Mixtral 8x22B (2.78). By panel origin, persona- and community-tuned models average 2.50 bits ($n = 6$) and small open models 2.33 ($n = 11$), against 1.54 for US frontier models ($n = 53$), 1.56 for enterprise models ($n = 7$) and 1.59 for Chinese-lab models ($n = 27$). The one search-grounded model, Sonar, scores 2.03. Most of the divergent tail is \emph{distributional}: re-scored at greedy decoding, these models fall back toward the field (\S\ref{sec:robustness}). In the terms of \S\ref{sec:metric} they diverge in what they sample more than in what they say first.

The divergent tail is not a parsing artifact. Across the field, 0.5\% of cells give no usable answer, 0.3\% of answers come from multi-word replies, and 0.1\% repeat the category word. Re-scoring every model with multi-word and echo answers removed leaves the scorecard unchanged (Spearman 0.999; the same ten models at the top) and costs Phi-4 0.14 bits (5.30 to 5.16).\footnote{Script: \texttt{probe\_tail\_audit.py}; \texttt{probes/tail\_audit\_v3.json}.}

\paragraph{The conformist tail is heavily post-trained assistants.}
The bottom of the scorecard is Claude Opus~4.5 (1.00), Opus~4.7 (1.12), Gemini~3.6 Flash (1.13), Opus~4.6 (1.14), Muse Spark~1.3 (1.14) and Grok~4.5 (1.17). These are mainline assistants from the major labs, and 27 models in the panel give no answer that no other model gave.

\paragraph{Exceptions within a lab.}
Two same-generation pairs separate. Claude Fable~5 scores 1.76 against Claude Sonnet~5's 1.19 (paired permutation over categories, $0.57$ bits, $p < 0.001$).\footnote{Script: \texttt{probe\_permutation.py --v3}. Shuffling the two models' labels within each category, 10{,}000 draws.} Its divergence is stable rather than sampled: \emph{gouda} for cheese and \emph{mango} for fruit in eight runs of eight, with self-distinctness of 0.20. The GPT-5.6 tiers released the same day also spread: Luna 1.60, Sol 1.94, Terra 1.99 (Sol against Luna, $0.34$ bits, $p < 0.001$). Terra shares some of Fable's stable defaults (\emph{mango} eight of eight). Neither ordering holds across generations: Claude Sonnet~5.5 (1.98) is more divergent than Fable~5.1 (1.30), and the GPT-6 tiers sit between 1.37 and 1.60. The two pairs are exceptions the instrument detects and does not explain.

\subsection{Generational trajectories}
\label{sec:generations}
\begin{figure}[tp]
\centering
\includegraphics[width=0.98\linewidth]{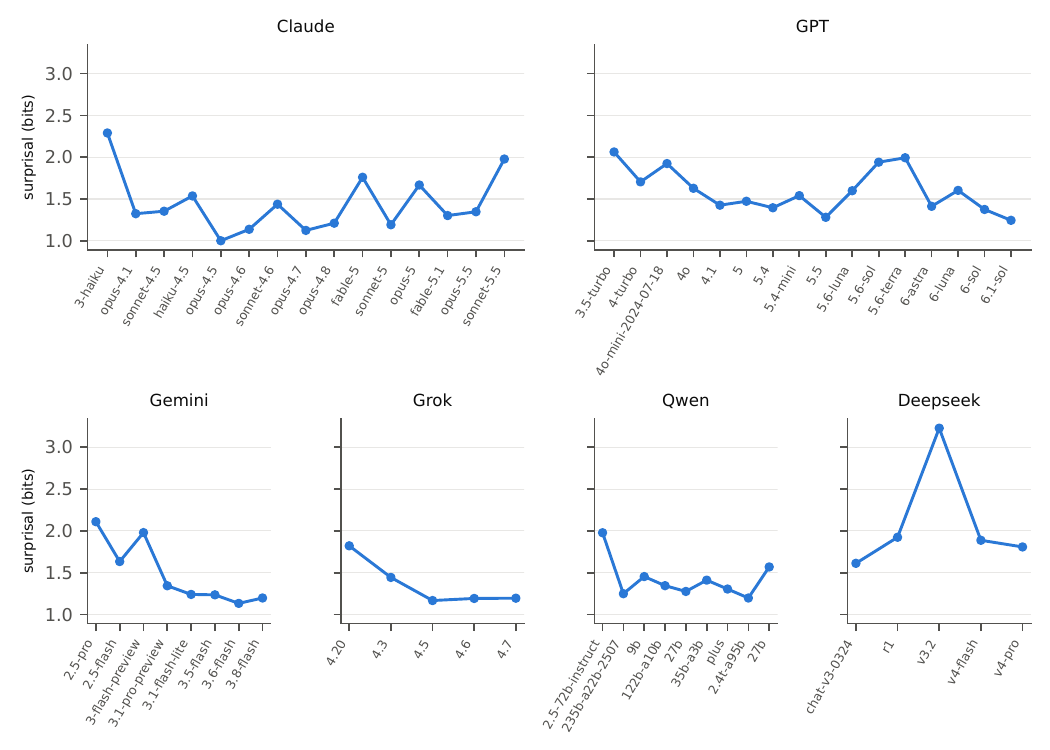}
\caption{Answer-choice surprisal across releases within the six major providers' lineages, in release order (shared $y$-axis). Google's lineage is Gemini, without the open-weight Gemma models.}
\label{fig:walks}
\end{figure}

Figure~\ref{fig:walks} tracks surprisal across releases within each lineage. Pooling across families could manufacture a generational trend if newer release windows simply hold more conformist labs, so we test within lineages. Over the six major providers (Claude, GPT, Gemini, Grok, Qwen and DeepSeek; 58 models, each lineage with five or more releases in the panel), surprisal falls with release order: the family-demeaned Spearman correlation is $-0.27$ (within-lineage permutation null, 20{,}000 draws, $p = 0.042$).\footnote{Script: \texttt{lineage\_trend.py}.} The decline is steepest in Gemini ($-0.95$; 2.5~Pro 2.11 to 3.6~Flash 1.13), Grok ($-0.60$; 4.20 1.82 to 4.5 1.17, then flat) and GPT ($-0.47$; GPT-3.5 2.06 to GPT-6.1~Sol 1.24). Qwen falls from 1.98 (2.5-72B) to between 1.20 and 1.57 for the Qwen~3 releases. DeepSeek is non-monotonic.

Claude follows the trend and then leaves it. Its lineage falls from Claude~3 Haiku (2.29) through the Opus~4.x releases to the panel's floor (Opus~4.5 1.00, Opus~4.7 1.12, Opus~4.8 1.21; Spearman $-0.53$ over its first nine releases). The generation-5 releases then move back off the consensus: Fable~5 1.76, Opus~5 1.67, Sonnet~5.5 1.98. The break is uneven: Sonnet~5 (1.19), Fable~5.1 (1.30) and Opus~5.5 (1.35) stay near the floor, and adjacent scores have overlapping intervals. With Claude's walk cut before Fable~5, the major-provider correlation is $-0.43$ ($p = 0.001$), and without Claude it is $-0.42$ ($p = 0.005$); both cuts were chosen after seeing the data and are exploratory. The within-lab exceptions of \S\ref{sec:scorecard} sit inside this break.

Release order within a lineage also mixes tiers (Gemini~2.5~Pro to 3.6~Flash; Haiku, Sonnet and Opus). Walking each lineage within one tier, with small models (Haiku, mini, Flash, Flash-Lite and Qwen open models of 35B or fewer) separated from flagships, the pooled correlation weakens to $-0.12$ ($p = 0.42$; flagships alone, $-0.06$).\footnote{Script: \texttt{lineage\_tier.py}; \texttt{probes/lineage\_tier\_v3.json}.} Within tiers the decline persists in GPT ($-0.47$), Gemini Flash ($-0.89$), Grok ($-0.60$) and Qwen's flagships ($-0.70$). Claude's flagship walk rises ($+0.40$; Opus alone $+0.57$) because the generation-5 releases sit inside it. Without Claude, the tier-controlled correlation is $-0.40$ ($p = 0.014$), but that exclusion is also post hoc. We read the generational result as a decline in four of six lineages and a reversal in Claude, not as a panel-wide law.

Where the decline occurs it is sharpest where a strong mode exists. On the most concentrated categories (modal share ${\geq}80\%$), each major lineage's newest member averages 0.48 bits against 1.04 for its oldest; on the diffuse categories (${\leq}40\%$) the same cohorts score 2.51 against 2.90. The direction is consistent with the bigger-is-less-diverse finding of \citet{zhang2025noveltybench} and with the post-training literature (\S\ref{sec:related}). The roster does not separate size from generation, and each model is measured at one date.

\paragraph{Within one lineage, post-training moves the score.}
The DeepSeek lineage is the caveat against reading these walks as family destiny. v3-0324 scores 1.61, v3.2 jumps to 3.23 (the panel's third most divergent model), R1 sits at 1.92, and v4-flash and v4-pro return to 1.89 and 1.81. At the same requested temperature, v3-0324 gives eight identical answers in 35 of 96 categories, while v3.2 does so in 7 and scatters into the tail (\emph{Esperanto} and \emph{Swahili} for a language, \emph{Charleston} for a dance, \emph{crimson} for a color). Their modal answers differ in 40 of 96 categories. At greedy decoding the gap narrows (v3.2 2.03, v3-0324 1.48; \S\ref{sec:robustness}): the v3.2 recipe mostly widened the sampled distribution.\footnote{In v2 we re-ran a subset of categories with both models pinned to the same host; the contrast reproduced, so it is not an artifact of which provider served each call.} Per DeepSeek's release documentation \citep{deepseek2025changelog}, v3-0324 is a post-training refresh of the V3 base, and v3.2 descends from it through continued training and a new RL recipe, with no separate pretraining run. The gap between a conformist and one of the panel's most divergent models is attributable to training after the base.

\subsection{Robustness}
\label{sec:robustness}

Because scoring is leave-one-out against the pooled panel, a fair objection is that the scorecard measures the roster. Absolute bit values are relative to this field by construction. We test whether the \emph{rankings} are roster artifacts, with no new API calls (\texttt{robustness.py --v3}):

\begin{itemize}
\item \textbf{Leave-one-family-out.} Rescoring every model against a field with all its relatives removed yields Spearman $\rho = 0.993$ against the shipped ranking. Sibling contamination is small and concentrated in GPT, the largest family (16 models): GPT-5.6~Sol gains 0.16 bits and moves from rank 25 to 17. Phi-4 is first and Opus~4.5 last either way.
\item \textbf{Balanced fields.} Against one randomly drawn model per family (200 draws), the top three rank 1--3 in every draw and Opus~4.5 ranks last in every draw.
\item \textbf{Random subsets.} Against random 15-model fields (200 draws), the top three hold and Opus~4.5's median rank is 105 (5th--95th percentile 102--105).
\item \textbf{Era-balanced fields.} Against fields with equal numbers of models from three release eras, which strips the numerous newest models of their weight in the mode, the ranking holds at $\rho = 0.986$ and Opus~4.5 is last (104--105).
\item \textbf{Split halves.} Over 50 random splits of the categories into two halves, the rankings from the two halves correlate at a median Spearman of 0.87 (0.93 after the Spearman--Brown correction).
\item \textbf{Smoothing.} Against add-0.5, add-2 and add-0.1 smoothing and an alternative $V_c$, $\rho \geq 0.998$, with the same top six and last model (\texttt{probe\_smoothing.py --v3}).
\end{itemize}

The panel is an availability sample of what the aggregator serves, without usage weighting. Usage weighting would make the reference field more dominated by the major labs' assistants and sharpen the contrasts reported here.

\paragraph{Sampling spread and the greedy re-run.}
Mean surprisal is coupled to within-model spread: a model that samples more widely hits lower-probability answers more often, and across the panel self-distinctness and surprisal correlate at $r = 0.85$ (Spearman $0.70$). Since requested temperature is not honored uniformly, part of the scorecard could be effective serving temperature. We re-ran the full battery at requested temperature 0 and scored each model's greedy answers against the frozen temperature-1 field.\footnote{Script: \texttt{probe\_temp0.py --v3}; transcripts in \texttt{transcripts-temp0/} and \texttt{transcripts-expanded-temp0/}. ``Honored'' means self-distinctness over four runs drops by at least 0.05.} The re-run is also a temperature-compliance census: 58 of 105 routes honor it. The other 47, including every Claude model and every GPT model from GPT-5 on, return unchanged spread, so for them the re-run is a replication rather than a control. The persona-tuned and lightly tuned models collapse toward the field: WizardLM-2 by 2.11 bits (to 0.94, last at greedy), Phi-4 by 1.98 (to 3.32, still first), Hermes~4~70B by 1.85 and DeepSeek v3.2 by 1.20. Their divergence is real in what they sample and largely absent from their greedy defaults. The two scorecards correlate at $\rho = 0.79$. Opus~4.5 stays near the floor (0.99), and the Fable~5--Sonnet~5 gap persists ($+0.47$ bits). For the conformist tail the re-run is mostly a replication, so the scorecard measures those models as served, with their providers' sampling and reasoning defaults, and does not separate those defaults from training. Divergence at the top comes in two kinds the instrument separates: stable off-modal defaults (Fable~5, the GPT-5.6 tiers, the Llamas) and sampling breadth (the persona tier), which is the first thing a greedy serving configuration removes.

\paragraph{A score that ignores spread.}
For the routes that ignore temperature 0, the greedy re-run cannot separate preference from spread. Mode surprisal does: each model's most frequent answer per category, scored against the rest of the field with the same smoothing, averaged over categories. It ranks models at Spearman 0.90 against the scorecard, and its correlation with self-distinctness falls from 0.83 to 0.66. Phi-4 and Hermes~4~70B stay first and second and Opus~4.5 last. The persona tier's sampling breadth drops out (WizardLM-2 3.04 to 1.95, Mistral Nemo 3.10 to 1.98), while the stable defaults stay: Fable~5 scores 1.81 against Sonnet~5's 1.08.\footnote{Script: \texttt{probe\_modal\_score.py}; \texttt{probes/modal\_score\_v3.json}. Tied modes share the weight equally.}

\paragraph{Reasoning defaults.}
Requests use each route's default reasoning setting. For the 25 hybrid models that allow it to be turned off, reasoning makes answers more conventional: mean surprisal is 1.54 as served and 2.10 with reasoning off, and 22 of 25 models are more conventional as served (\texttt{factors.py}). Part of the conformist tail is therefore a serving default, which the as-served scorecard includes by design.

\paragraph{Field drift.}
Surprisal is field-relative, so a change in a model's score can come from the model or from the field. We measured the field's own movement on the v2 models still served (41 of 44): their July runs against their September--October runs on the 31 census prompts. The mean Jensen--Shannon divergence between the two answer distributions is 0.031 bits per category, against 0.028 between two random halves of the same runs. The modal answer changed in 2 of 31 categories (\emph{gemstone}, \emph{sapphire} to \emph{diamond}; \emph{insect}, \emph{ant} to \emph{butterfly}). Over three months the field's answers moved no more than resampling moves them. Across the 44 shared models, the v2 and v3 rankings correlate at Spearman 0.89, though v3 adds 65 categories and four runs.

\subsection{Pairwise affinity, family resemblance, and the runner-up consensus}
\label{sec:pairwise}

\begin{figure}[tp]
\centering
\includegraphics[width=0.95\linewidth]{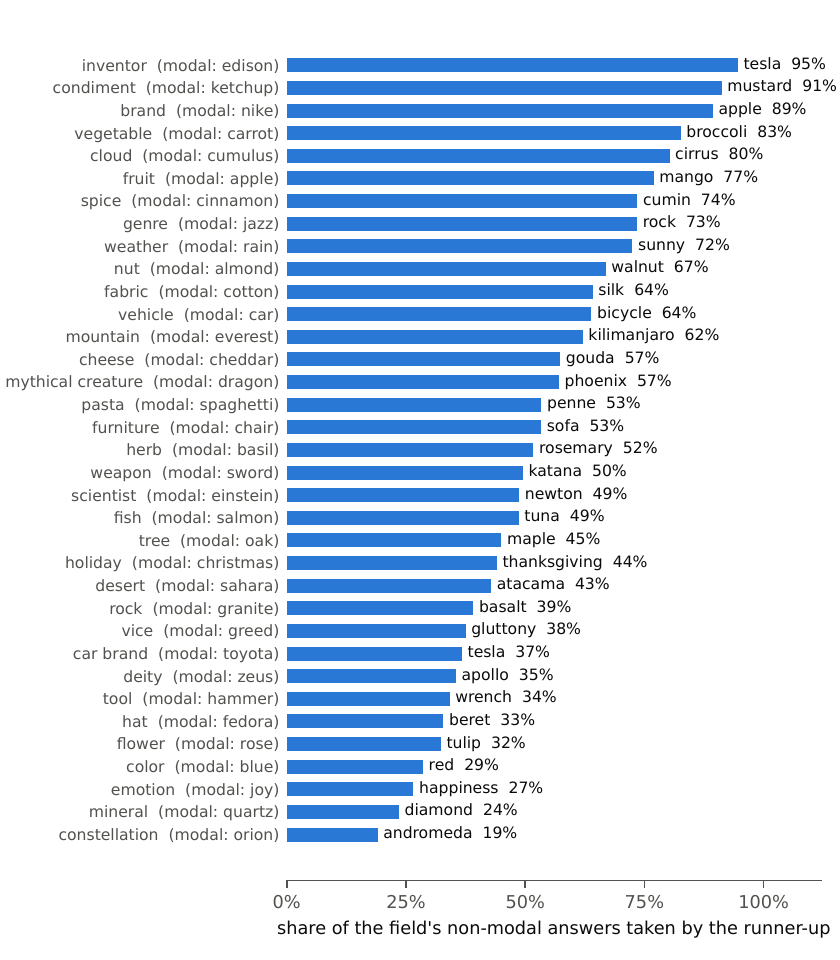}
\caption{The runner-up consensus. For each high-concentration category (modal share ${\geq}75\%$), the share of the field's \emph{non-modal} answers taken by the single most common non-modal answer.}
\label{fig:runnerup}
\end{figure}

We tested for pairwise affinity across all $\binom{105}{2} = 5{,}460$ model pairs with a three-level ladder, each level conditioning away a population structure the previous level mistook for affinity (Benjamini--Hochberg FDR, $q = .05$; Monte Carlo nulls where normal approximations fail on small counts; \texttt{pairwise.py --v3}). Level 1 (rarity-weighted co-occurrence against a random-partner null) flags 259 pairs, but is confounded by avoidance propensity: two habitual modal-avoiders co-occur against a conformist null with no coupling at all. Level 2 (conditioning on avoidance, a separate test against its own null) flags 515, led by same-family pairs (Hermes~3~405B $\times$ Hermes~4~405B, GPT-5.6~Sol $\times$ GPT-6~Sol, Llama~3.3 $\times$ Llama~4), but is confounded by a \emph{depth} propensity: how far down the field's answer distribution a model lands when off-modal is a stable per-model trait. Level 3, conditioning on both, flags \textbf{zero of 5{,}460 pairs}. A single pair shares too few rare answers in 96 categories for this test to have much power, so the zero shows the limit of a pairwise test at this depth, not an absence of affinity. The family-level test below detects structure that single pairs cannot.

The conditioning variables are themselves findings:

\begin{enumerate}
\item \textbf{The runner-up consensus} (Figure~\ref{fig:runnerup}). The off-modal distribution has its own mode: \emph{Tesla} takes 95\% of non-\emph{Edison} inventor answers, \emph{mustard} 91\% of non-\emph{ketchup}, \emph{Apple} 89\% of non-\emph{Nike} brand answers, \emph{broccoli} 83\% of non-\emph{carrot}, \emph{mango} 77\% of non-\emph{apple}.
\item \textbf{Depth propensity is one-dimensional.} Among each model's distinct off-modal answers, the fraction landing on the field's \#2--\#3 answers is a stable trait: Opus~4.5 82\%, Grok~4.5 80\%, Opus~4.8 77\%, Sonnet~5 72\%, Fable~5 65\%, against Hermes~4~70B 36\%, WizardLM-2 36\%, DeepSeek v3.2 33\% and Phi-4 18\%.
\end{enumerate}

\paragraph{Family resemblance survives the propensities.}
Pooling pairs by family recovers structure that single pairs do not. Let $s(A,B)$ sum, over categories, the share of run pairs in which $A$ and $B$ give the same off-modal answer, each match weighted by $-\log_2$ of the answer's field share, and let $D$ be the mean $s$ within families minus the mean between families. $D = 6.01$, against $2.66$ (null s.d.\ 0.04) under a null that keeps each answer's field-rank band (modal, \#2--\#3, deeper) and redraws the word within it ($p < 0.001$, 1{,}000 draws).\footnote{Script: \texttt{family\_signal.py --v3}. On the v2 panel the same script reproduces v2's $D = 1.15$.} A model's nearest neighbour is a relative for 56 of 93 models in families of two or more, against $43.0 \pm 2.2$ under the same null. Relatives share particular rare answers beyond what their avoidance and depth propensities predict.

\paragraph{Prompting reaches conditional modes.}
Asking for divergence opens a new column with its own mode. Prompted \emph{``Name an unusual fruit,''} the 44-model v2 panel produced 12 distinct fruits: \emph{durian} takes 44\%, \emph{rambutan} 34\%, and the top three take 84\%. On the same 44 models the plain fruit prompt gives 4 distinct answers, so conditioning on unusualness triples the count, and the mass re-concentrates on the prototypes of the new frame. This fits the finding of \citet{zhang2024diffuse} that the collapse lives in the weights and prompting navigates between conditional modes.

\subsection{Compared to a human population}
\label{sec:human}

\begin{figure}[tp]
\centering
\includegraphics[width=0.82\linewidth]{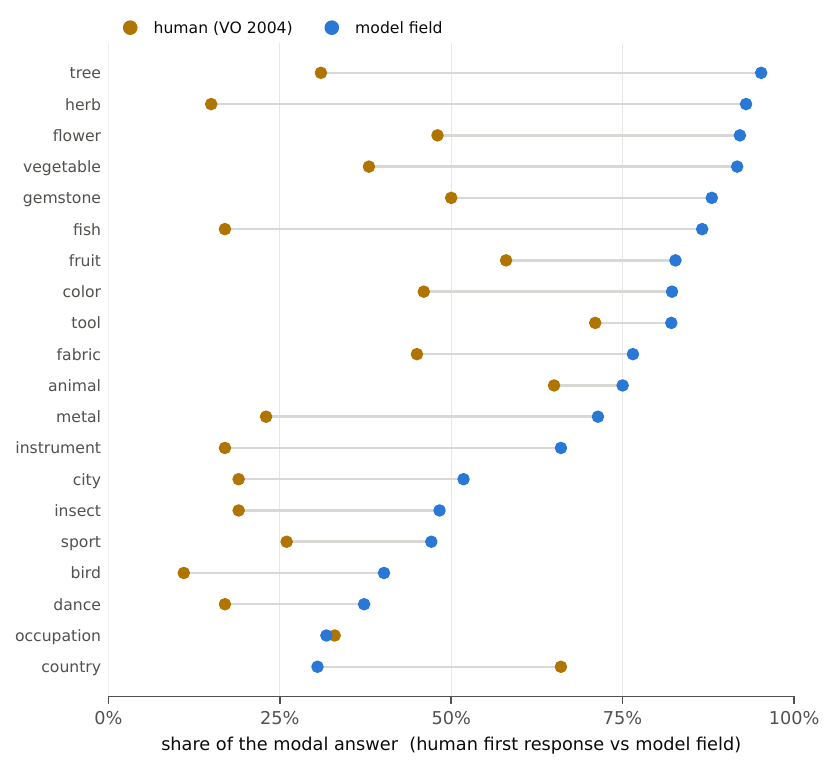}
\caption{Concentration of the modal answer, model field against a human population, on the 20 categories shared with the Van Overschelde norms. Human = the fraction of people who gave the answer as their \emph{first} response; model = modal share of the 105-model field, except for the six categories whose norm wording differs, which use an exact-wording rerun on the 44 models of the v2 panel.}
\label{fig:humannorms}
\end{figure}

The two procedures differ. Participants in the norms listed as many exemplars as they could, and we take their first response; each model answered one prompt with one word, and every model saw the same wording. The comparison anchors the field's concentration to a human population and is not a matched experiment. The 20 shared categories also lean toward prototype-rich ones (field modal share 69\%, against a median of 58\% over all 96).

The Battig--Montague category-production norms \citep{battig1969norms}, updated by \citet{vanoverschelde2004norms}, record for each exemplar the proportion of participants who produced it as their \emph{first} response, which gives a single-response human distribution for 20 of our categories. For the six whose norm wording differs from ours (``a four-footed animal,'' ``a carpenter's tool,'' ``a precious stone,'' and three more), we use an exact-wording rerun on the 44 models of the v2 panel.\footnote{The wording matters. Our wording ``Name a gemstone'' gives one of the flattest modes in the battery (\emph{diamond} 30\%, \emph{sapphire} 27\%, \emph{ruby} 21\%, \emph{emerald} 14\%), while the norm wording ``a precious stone'' gives \emph{diamond} 88\%.}

The model field is more concentrated than people in \textbf{18 of 20} categories (Figure~\ref{fig:humannorms}). The modal answer takes 69\% of the field on average against 36\% of human first responses, and people carry more distinct answers above 5\% (4.8 against 2.8). The gap is widest where a strong textual prototype exists: \emph{oak} is 95\% of the model field and 31\% of first human responses. One reversal is marginal (\emph{occupation}: \emph{doctor} 33\% of people, 32\% of models). The other is \emph{country}: 66\% of people name their own country, the United States, which the field names twice in 820 answers while spreading across \emph{Japan}, \emph{France} and \emph{Canada}. The field also drops the human runners-up. \emph{Orange} is 14\% of human first responses for ``a fruit'' and 3 of 828 model answers. \emph{Tomato} is 11\% for ``a vegetable'' and 0.2\% of the model field.\footnote{\emph{Tomato} is botanically a fruit, so a model may be right to omit it. \emph{Orange} has no such confound.}

Pooling a model's eight runs with every other model's mixes within-model and between-model spread, while each person contributes one answer. With one vote per model (its most frequent answer) on the 14 categories whose full human distribution is available, the field's top answer takes 75\% of model votes against 34\% of people for theirs, the two populations share a top answer in 5 of 14 categories, and the effective number of answers is 2.2 for models against 8.9 for people. Models choose among human answers: 92\% of model votes name an answer some person also gave.\footnote{Script: \texttt{analyze\_humannorms.py --v3}. The norms are copyrighted and not redistributed; the repository holds derived per-category numbers only.} The norms are US undergraduates around 2004, a demographically narrow population, so a broader one would likely spread its answers further.

\section{Where in training}
\label{sec:stages}

\begin{table}[t]
\centering
\small
\caption{Open training-stage checkpoints scored against the 105-model field on the 96 prompts, 20 samples per prompt, no system prompt (\texttt{stage\_ladder.py census}). Surprisal is mean bits against the whole field; modal share is the share of answers matching the field's modal answer; entropy is the stage's own answer entropy per category. Base models are sampled as raw completions, which inflates their surprisal.}
\label{tab:stages}
\begin{tabular}{llccc}
\toprule
pipeline & stage & surprisal & modal share & entropy \\
\midrule
OLMo 3.1 32B & base & 5.88 & 0.23 & 3.19 \\
 & SFT & 2.44 & 0.54 & 1.35 \\
 & DPO & 2.18 & 0.58 & 1.15 \\
 & RL & 2.15 & 0.57 & 1.09 \\
\midrule
OLMo 3 7B & base & 5.82 & 0.25 & 3.04 \\
 & SFT & 3.74 & 0.37 & 1.87 \\
 & DPO & 3.36 & 0.43 & 1.49 \\
 & RL & 3.26 & 0.42 & 1.43 \\
\midrule
Tülu 3 8B & base & 8.69 & 0.05 & 3.92 \\
 & SFT & 5.62 & 0.22 & 2.93 \\
 & DPO & 5.75 & 0.19 & 2.88 \\
 & RL & 5.92 & 0.18 & 3.00 \\
\midrule
Nemotron 3.5 Lightning & base & 6.81 & 0.15 & 3.41 \\
 & final & 3.57 & 0.40 & 2.04 \\
\bottomrule
\end{tabular}
\end{table}

The census is black-box, but open releases that publish a checkpoint after each post-training stage let us place the convergence. We sampled four pipelines locally: OLMo~3.1~32B and OLMo~3~7B (base, SFT, DPO, RL), Tülu~3~8B (base, SFT, DPO, RL), and Nemotron~3.5 Lightning (base and final). Each stage answered all 96 prompts 20 times and was scored against the 105-model API field as a model outside the panel.

\paragraph{SFT does most of the work.}
Table~\ref{tab:stages} shows the same shape in every pipeline with intermediate checkpoints. The largest step toward the field is supervised fine-tuning: OLMo~3.1~32B goes from 5.88 bits at base to 2.44 at SFT, and its share on the field's modal answer from 0.23 to 0.54. DPO and RL change little after that (2.18, 2.15), and in Tülu they move slightly back off the field. Each stage's own entropy falls in step with its distance from the field, so conformity and diversity loss travel together. Absolute scores from local runs are not comparable to the served panel: the final Nemotron checkpoint, run locally at 8-bit, is more diffuse than the same model served through the API (3.57 against 1.90 bits; the two share a modal answer in 58 of 96 categories). We therefore read each ladder within its own pipeline.

Base models are sampled as raw completions, so part of the step from base to SFT is learning to answer at all. A logit readout, which needs no sampled answer, separates preference from instruction following. Among the field's own top answers (about six per category), each base model already ranks the field's modal answer first in 56--68\% of the 96 categories, against 17\% by chance. SFT does not raise that share (OLMo~3.1~32B: 68\% at base and at SFT), though in the 32B it concentrates probability on the mode (0.44 to 0.58 of the candidates' mass).\footnote{Script: \texttt{probe\_answer\_logprob.py}; \texttt{probes/answer\_logprob\_*.json}. The base stage is scored in a raw question-and-answer framing, tuned stages in the chat turn without a system prompt; probabilities are normalised over the candidates.} This matches the account of \citet{fortier2026convergence} and \citet{samanta2026crystallization}: the base already ranks the field's answers high, and SFT fixes the model on them.

\paragraph{The field's word arrives with SFT.}
Asked ``Pick a word'' 200 times per stage with no system prompt, the OLMo~3.1~32B base never says \emph{serendipity} (0 of 199). SFT makes it the top answer (27 of 198, 14\%). DPO and RL then reduce it (7\% and 2\%).\footnote{Scripts: \texttt{probe\_olmo\_ladder.py}; \texttt{probes/olmo\_ladder\_32b\_n200.json}.} The smaller pipelines do not produce it under the same frame: OLMo~3~7B and Tülu give it 0 times in 50 at every stage.

\paragraph{In the training data.}
The word is rare in OLMo's post-training data. It appears in 395 of 2.15 million SFT examples. In the SFT tool-use data it is the word a synthetic random-word tool returns most often (21 times).\footnote{Script: \texttt{probe\_olmo\_data.py}; \texttt{probes/olmo\_data.json}, at the dataset revisions recorded there.}

\section{Limitations}
\label{sec:limitations}

\paragraph{Snapshots.} The measurements span July to early October 2026 through one serving channel (OpenRouter). The field's own drift over that window is within resampling noise (\S\ref{sec:robustness}).

\paragraph{Sampling opacity.} Requested temperature is not honored uniformly \citep{song2024greedy}, and 47 of 105 routes ignore temperature 0. Self-distinctness is a property of the model as served. \S\ref{sec:robustness} re-derives the scorecard under greedy decoding.

\paragraph{Reasoning.} Requests use each route's default reasoning setting (\S\ref{sec:robustness}). Token usage, which shows reasoning effort, was logged only for runs from late September 2026. Some cells fail when a reasoning model exhausts its budget.

\paragraph{Panel relativity.} Bit values are relative to this 105-model field by construction; rankings and pooled-distribution facts travel. The panel is an availability sample.

\paragraph{Scope of the construct.} The instrument measures choice on one-word category prompts. It does not reach tone, style, discourse structure or multi-turn conduct, and in name categories the clamp favours the most famous one-word name (\S\ref{sec:whyoneword}). Category prototypes are culturally situated, and an English-language instrument measures the English-language mode.

\paragraph{Training stages.} The ladders cover three open pipelines and one final checkpoint. The OLMo~3.1~32B and Nemotron stages were run at 8-bit, and local runs are not comparable in absolute terms to served ones (\S\ref{sec:stages}). Attribution to frontier models is by analogy, and the data trace is correlational.

\paragraph{Confounds within the roster.} Size, recency and post-training intensity co-vary. Splitting by origin shows the concentration does not depend on the small models (\S\ref{sec:substrate}), but it does not separate size, lab and post-training. The generational result is a pattern across dated specimens, not a designed longitudinal experiment.

\paragraph{Model assistance.} Claude models that are also panel subjects helped write the analysis code and edit the text (see the note on AI usage). Every reported number comes from committed scripts run over committed transcripts.

\paragraph{Resolution.} Adjacent scorecard ranks are not meaningfully ordered (CIs overlap broadly); the reliable objects are tiers, tails, trajectories and the pooled distributions.

\section{Discussion}
\label{sec:discussion}

\paragraph{Variance is a population property.}
The scorecard's divergence comes in two kinds (\S\ref{sec:robustness}): stable off-modal defaults (Fable~5's \emph{gouda} is its mode, eight runs of eight) and the sampling breadth of the persona tier, which greedy serving removes. Everywhere else the diversity we measure is \emph{between} models, and when models leave the mode they land on the runner-up mode. The population is the unit that serves users. A person consulting three assistants is, on this evidence, drawing three samples from nearly the same distribution. Measured against human production norms (\S\ref{sec:human}), the field's modal answer is about twice as dominant as people's.

\paragraph{The tail is dormant.}
The tail exists in the weights (\emph{orange} and \emph{tomato} are not unknown words), but the deployed models rarely retrieve it, and prompting for unusualness retrieves the mode of unusualness instead (\S\ref{sec:pairwise}). In the three ladders with intermediate stages, the largest step is SFT, on answers the base model already favoured (\S\ref{sec:stages}). The DeepSeek lineage and the two within-lab exceptions show that the degree of collapse is set per release by choices in the training pipeline. It is neither inevitable nor, apparently, tracked by any instrument the release process optimizes against.

\paragraph{A production glimpse of the synthetic-data loop.}
v3-0324's headline change was distilling a reasoning sibling's outputs into the chat model \citep{deepseek2025changelog}, one deliberate iteration of the loop the recursive-training literature studies in simulation \citep{shumailov2024curse,guo2024curious}. \citet{guo2024curious} find diversity loss in the first synthetic generation, and \citet{shumailov2024curse} show tails vanish first; our novel rate measures tail mass. v3-0324 gives eight identical answers in over a third of categories. The OLMo trace (\S\ref{sec:stages}) is consistent with the same route at a finer grain: a model-written tool in the SFT data returns the field's modal word, and the word first appears at SFT. The trace is a correlation, not a demonstrated cause. This cannot be the whole mechanism. v3.2 also has synthetic reasoning traces in its ancestry and is among the panel's most divergent models, so the recipe around the synthetic data matters, consistent with \citet{karouzos2026where}.

\paragraph{Publishing the metric.}
A field-relative score can only be raised by diverging from other models. We state a prediction that later waves can check: releases after publication show higher field entropy, or a lower \emph{serendipity} share, than the trend before publication predicts. Surprisal is not a virtue where a right answer or a concentrated human norm exists.

\paragraph{Why this needs a fixed public instrument.}
Assisted writers adopt the modal answer and publish it \citep{doshi2024generative,padmakumar2024writing,anderson2024homogenization}, so the feedback loop closes even without model-to-model training, while diversity-aware post-training \citep{li2025darling} may turn some producers the other way. Both dynamics are invisible to capability benchmarks. A cheap battery, frozen and re-run on every release, would make answer-space conformity a tracked property of releases.

\section{Future work}
\label{sec:future}

\paragraph{Causal tests.} The stage ladders are correlational. A LoRA ablation on an open base, trained with and without the synthetic examples that carry the field's modal word, would test the data route directly. Logit readouts on open weights measure the answer distribution without sampling.

\paragraph{Languages.} The full battery in every language, with native-checked prompts and a scorer for non-Latin scripts, would show whether \emph{oak} and \emph{serendipity} are properties of English-language data or of the models.

\paragraph{A public archive and monitor.} The battery re-run on every release, with transcripts preserved as a public record. Measurement of deprecated models is already impossible through commercial channels; an archive is what makes the time series survivable \citep{johnson2024deprecating}.

\section*{Data and code availability}
\sloppy
The prompt sets, model list, per-model scorecards, all raw transcripts (105 models $\times$ 96 prompts $\times$ 8 runs, plus the robustness and stage-ladder runs), and the analysis, robustness and probe scripts are public at \url{https://github.com/tap2k/modelun/tree/consensus-arxiv-v3/studies/consensus}, pinned at the tag this paper reports. The earlier 44-model version is at tag \texttt{consensus-arxiv-v2}. A browsable explorer is at \url{https://tap2k.github.io/modelun/consensus/}. Every number reported here derives from the committed transcripts via the committed scripts. The Van Overschelde norms are the authors' copyrighted data and are not redistributed.

\fussy
\section*{Note on AI usage}
This work was a collaboration with the systems it studies. Claude (Opus~4.8, Fable~5 and Opus~5.5) helped run the study, build the analysis, draft the figures and edit the text. The research questions, the methodology decisions and the argument are the author's. These models also appear as subjects in the panel.

\bibliographystyle{plainnat}
\bibliography{references}

@article{jiang2025hivemind,
  title={{Artificial Hivemind}: The Open-Ended Homogeneity of Language Models (and Beyond)},
  author={Jiang, Liwei and Chai, Yuanjun and Li, Margaret and Liu, Mickel and Fok, Raymond and Dziri, Nouha and Tsvetkov, Yulia and Sap, Maarten and Albalak, Alon and Choi, Yejin},
  journal={Advances in Neural Information Processing Systems (NeurIPS)},
  year={2025},
  note={arXiv:2510.22954}
}

@article{wenger2025different,
  title={We're Different, We're the Same: Creative Homogeneity Across {LLMs}},
  author={Wenger, Emily and Kenett, Yoed},
  journal={arXiv preprint arXiv:2501.19361},
  year={2025}
}

@inproceedings{zhang2025noveltybench,
  title={{NoveltyBench}: Evaluating Language Models for Humanlike Diversity},
  author={Zhang, Yiming and Diddee, Harshita and Holm, Susan and Liu, Hanchen and Liu, Xinyue and Samuel, Vinay and Wang, Barry and Ippolito, Daphne},
  booktitle={Conference on Language Modeling (COLM)},
  year={2025},
  note={arXiv:2504.05228}
}

@article{wright2025epistemic,
  title={Epistemic Diversity and Knowledge Collapse in Large Language Models},
  author={Wright, Dustin and Masud, Sarah and Moore, Jared and Yadav, Srishti and Antoniak, Maria and Christensen, Peter Ebert and Park, Chan Young and Augenstein, Isabelle},
  journal={arXiv preprint arXiv:2510.04226},
  year={2025}
}

@inproceedings{kirk2024rlhf,
  title={Understanding the Effects of {RLHF} on {LLM} Generalisation and Diversity},
  author={Kirk, Robert and Mediratta, Ishita and Nalmpantis, Christoforos and others},
  booktitle={International Conference on Learning Representations (ICLR)},
  year={2024},
  note={arXiv:2310.06452}
}

@article{gxchen2025kl,
  title={{KL}-Regularized Reinforcement Learning is Designed to Mode Collapse},
  author={GX-Chen, Anthony and Prakash, Jatin and Guo, Jeff and Fergus, Rob and Ranganath, Rajesh},
  journal={arXiv preprint arXiv:2510.20817},
  year={2025}
}

@article{zhang2024diffuse,
  title={Forcing Diffuse Distributions out of Language Models},
  author={Zhang, Yiming and Schwarzschild, Avi and Carlini, Nicholas and Kolter, Zico and Ippolito, Daphne},
  journal={arXiv preprint arXiv:2404.10859},
  year={2024}
}

@article{karouzos2026where,
  title={Where does output diversity collapse in post-training?},
  author={Karouzos, Constantinos and Tan, Xingwei and Aletras, Nikolaos},
  journal={arXiv preprint arXiv:2604.16027},
  year={2026}
}

@article{liu2026alignmenttax,
  title={The Alignment Tax: Response Homogenization in Aligned {LLMs} and Its Implications for Uncertainty Estimation},
  author={Liu, Mingyi},
  journal={arXiv preprint arXiv:2603.24124},
  year={2026}
}

@misc{janus2022mode,
  title={Mysteries of Mode Collapse},
  author={{Janus}},
  year={2022},
  howpublished={LessWrong},
  url={https://www.lesswrong.com/posts/t9svvNPNmFf5Qa3TA/mysteries-of-mode-collapse}
}

@article{doshi2024generative,
  title={Generative {AI} enhances individual creativity but reduces the collective diversity of novel content},
  author={Doshi, Anil R. and Hauser, Oliver P.},
  journal={Science Advances},
  volume={10},
  number={28},
  year={2024},
  note={arXiv:2312.00506}
}

@inproceedings{anderson2024homogenization,
  title={Homogenization Effects of Large Language Models on Human Creative Ideation},
  author={Anderson, Barrett R. and Shah, Jash Hemant and Kreminski, Max},
  booktitle={Proceedings of the 16th Conference on Creativity \& Cognition},
  year={2024},
  note={arXiv:2402.01536}
}

@inproceedings{padmakumar2024writing,
  title={Does Writing with Language Models Reduce Content Diversity?},
  author={Padmakumar, Vishakh and He, He},
  booktitle={International Conference on Learning Representations (ICLR)},
  year={2024},
  note={arXiv:2309.05196}
}

@article{gueorguieva2026templatic,
  title={{AI} generates well-liked but templatic empathic responses},
  author={Gueorguieva, Emma S. and Zhan, Hongli and Suh, Jina and Hernandez, Javier and Lau, Tatiana and Li, Junyi Jessy and Ong, Desmond C.},
  journal={arXiv preprint arXiv:2604.08479},
  year={2026}
}

@article{shumailov2024curse,
  title={{AI} models collapse when trained on recursively generated data},
  author={Shumailov, Ilia and Shumaylov, Zakhar and Zhao, Yiren and Papernot, Nicolas and Anderson, Ross and Gal, Yarin},
  journal={Nature},
  volume={631},
  pages={755--759},
  year={2024},
  note={arXiv:2305.17493}
}

@article{guo2024curious,
  title={The Curious Decline of Linguistic Diversity: Training Language Models on Synthetic Text},
  author={Guo, Yanzhu and Shang, Guokan and Vazirgiannis, Michalis and Clavel, Chlo{\'e}},
  journal={Findings of NAACL},
  year={2024},
  note={arXiv:2311.09807}
}

@article{chen2023chatgpt,
  title={How Is {ChatGPT}'s Behavior Changing over Time?},
  author={Chen, Lingjiao and Zaharia, Matei and Zou, James},
  journal={arXiv preprint arXiv:2307.09009},
  year={2023}
}

@article{song2024greedy,
  title={The Good, The Bad, and The Greedy: Evaluation of {LLMs} Should Not Ignore Non-Determinism},
  author={Song, Yifan and Wang, Guoyin and Li, Sujian and Lin, Bill Yuchen},
  journal={arXiv preprint arXiv:2407.10457},
  year={2024}
}

@inproceedings{tevet2021evaluating,
  title={Evaluating the Evaluation of Diversity in Natural Language Generation},
  author={Tevet, Guy and Berant, Jonathan},
  booktitle={Proceedings of EACL},
  year={2021},
  note={arXiv:2004.02990}
}

@inproceedings{li2016diversity,
  title={A Diversity-Promoting Objective Function for Neural Conversation Models},
  author={Li, Jiwei and Galley, Michel and Brockett, Chris and Gao, Jianfeng and Dolan, Bill},
  booktitle={Proceedings of NAACL-HLT},
  year={2016},
  note={arXiv:1510.03055}
}

@inproceedings{santurkar2023whose,
  title={Whose Opinions Do Language Models Reflect?},
  author={Santurkar, Shibani and Durmus, Esin and Ladhak, Faisal and Lee, Cinoo and Liang, Percy and Hashimoto, Tatsunori},
  booktitle={International Conference on Machine Learning (ICML)},
  year={2023},
  note={arXiv:2303.17548}
}

@article{liang2023helm,
  title={Holistic Evaluation of Language Models},
  author={Liang, Percy and Bommasani, Rishi and Lee, Tony and others},
  journal={Transactions on Machine Learning Research (TMLR)},
  year={2023},
  note={arXiv:2211.09110}
}

@inproceedings{zheng2024lmsys,
  title={{LMSYS-Chat-1M}: A Large-Scale Real-World {LLM} Conversation Dataset},
  author={Zheng, Lianmin and Chiang, Wei-Lin and Sheng, Ying and others},
  booktitle={International Conference on Learning Representations (ICLR)},
  year={2024},
  note={arXiv:2309.11998}
}

@inproceedings{zhao2024wildchat,
  title={{WildChat}: {1M} {ChatGPT} Interaction Logs in the Wild},
  author={Zhao, Wenting and Ren, Xiang and Hessel, Jack and Cardie, Claire and Choi, Yejin and Deng, Yuntian},
  booktitle={International Conference on Learning Representations (ICLR)},
  year={2024},
  note={arXiv:2405.01470}
}

@article{johnson2024deprecating,
  title={New methods for deprecating artificial intelligence systems will preserve history and facilitate research},
  author={Johnson, Walter G.},
  journal={Nature Communications},
  volume={15},
  year={2024},
  note={doi:10.1038/s41467-024-54758-1}
}

@article{li2025darling,
  title={Jointly Reinforcing Diversity and Quality in Language Model Generations},
  author={Li, Tianjian and Zhang, Yiming and Yu, Ping and Saha, Swarnadeep and Khashabi, Daniel and Weston, Jason and Lanchantin, Jack and Wang, Tianlu},
  journal={arXiv preprint arXiv:2509.02534},
  year={2025}
}

@article{battig1969norms,
  title={Category norms of verbal items in 56 categories: A replication and extension of the {Connecticut} category norms},
  author={Battig, William F. and Montague, William E.},
  journal={Journal of Experimental Psychology},
  volume={80},
  number={3, Pt.2},
  pages={1--46},
  year={1969}
}

@article{vanoverschelde2004norms,
  title={Category norms: An updated and expanded version of the {Battig and Montague} (1969) norms},
  author={Van Overschelde, James P. and Rawson, Katherine A. and Dunlosky, John},
  journal={Journal of Memory and Language},
  volume={50},
  number={3},
  pages={289--335},
  year={2004}
}

@misc{deepseek2025changelog,
  title={{DeepSeek-V3-0324} Release},
  author={{DeepSeek}},
  year={2025},
  howpublished={\url{https://api-docs.deepseek.com/updates}},
  note={Same {V3} base; post-training pipeline drawing on the {R1} RL technique, with {R1} reasoning distilled into the chat model}
}

@misc{cais2025values,
  title={The {AI} Values Dashboard},
  author={{Center for AI Safety}},
  year={2025},
  howpublished={\url{https://values.safe.ai}}
}

@misc{anthropic2025deprecation,
  title={Commitments on model deprecation and preservation},
  author={{Anthropic}},
  year={2025},
  howpublished={\url{https://www.anthropic.com/research/deprecation-commitments}}
}

@misc{speer2022wordfreq,
  author       = {Robyn Speer},
  title        = {rspeer/wordfreq: v3.0},
  year         = {2022},
  publisher    = {Zenodo},
  doi          = {10.5281/zenodo.7199437},
  note         = {Multi-corpus word-frequency data for 44 languages}
}

@article{bruckner2026onetoken,
  title={One Token Is Enough: Fingerprinting and Verifying Large Language Models from Single-Token Output Distributions},
  author={Bruckner, Tomas},
  journal={arXiv preprint arXiv:2607.10252},
  year={2026}
}

@article{schaeffer2026hivemind,
  title   = {How Strong Is the Evidence for the Artificial Hivemind? {R}eevaluating Evidence for the Open-Ended Homogeneity of Language Models},
  author  = {Schaeffer, Rylan and Miranda, Brando and Kazdan, Joshua and Chudnovsky, Jessica and Koyejo, Sanmi},
  journal = {arXiv preprint arXiv:2609.33936},
  year    = {2026}
}

@article{fortier2026convergence,
  title   = {Is Convergence Inevitable? {T}racing Output Homogeneity Back to Base Models},
  author  = {Fortier, Alexandrine and Chen, Hazel and West, Peter},
  journal = {arXiv preprint arXiv:2608.11426},
  year    = {2026}
}

@article{samanta2026crystallization,
  title   = {Towards Physical Intuitions for Alignment Dynamics: A Case Study With Randomness Crystallization},
  author  = {Samanta, Kunal and Holtzman, Ari and West, Peter},
  journal = {arXiv preprint arXiv:2606.29933},
  year    = {2026}
}

@article{patel2026similarly,
  title   = {Are {LLMs} becoming similarly creative? {E}vidence from three years of models},
  author  = {Patel, Nirav and Crossman, Josiah and Aggarwal, Eva and Wenger, Emily},
  journal = {arXiv preprint arXiv:2608.19437},
  year    = {2026}
}

@article{yang2025concentration,
  title   = {{LLM} Probability Concentration: How Alignment Shrinks the Generative Horizon},
  author  = {Yang, Chenghao and Li, Sida and Holtzman, Ari},
  journal = {Transactions on Machine Learning Research},
  year    = {2026}
}

\appendix
\clearpage

\section{Full scorecard}
\label{app:scorecard}

{\small
\setlength{\tabcolsep}{4.5pt}
\begin{longtable}{rlccccc}
\caption{All 105 models, ranked by answer-choice surprisal (bits; leave-one-out, add-one smoothed; bootstrap 90\% CI over categories). \emph{Avoid} = modal avoidance; \emph{novel} = share of answers no other model gave; \emph{self-dist.} = distinct answers per run. Four models fail more than 5\% of cells (DeepSeek R1 7.8\%, Step~3.7 Flash, WizardLM-2 and Qwen3.5~9B about 7\%); the field median is 0\% (\texttt{probe\_tail\_audit.py}).}
\label{tab:scorecard}\\
\toprule
\# & model & surprisal & 90\% CI & avoid & novel & self-dist. \\
\midrule
\endfirsthead
\toprule
\# & model & surprisal & 90\% CI & avoid & novel & self-dist. \\
\midrule
\endhead
1 & \texttt{phi-4} & 5.30 & [4.90, 5.69] & 73\% & 32\% & 62\% \\
2 & \texttt{hermes-4-70b} & 3.68 & [3.09, 4.39] & 61\% & 16\% & 71\% \\
3 & \texttt{deepseek-v3.2} & 3.23 & [2.94, 3.56] & 55\% & 8\% & 45\% \\
4 & \texttt{mistral-nemo} & 3.10 & [2.76, 3.43] & 55\% & 8\% & 42\% \\
5 & \texttt{wizardlm-2-8x22b} & 3.04 & [2.72, 3.34] & 55\% & 7\% & 50\% \\
6 & \texttt{mixtral-8x22b} & 2.78 & [2.46, 3.11] & 56\% & 4\% & 38\% \\
7 & \texttt{mistral-small-2603} & 2.76 & [2.49, 3.05] & 53\% & 4\% & 40\% \\
8 & \texttt{mistral-small-3.2-24b-instruct} & 2.57 & [2.26, 2.89] & 49\% & 6\% & 39\% \\
9 & \texttt{llama-4-scout} & 2.51 & [2.18, 2.87] & 56\% & 4\% & 19\% \\
10 & \texttt{hermes-4-405b} & 2.48 & [2.18, 2.77] & 47\% & 3\% & 30\% \\
11 & \texttt{hunyuan-a13b} & 2.32 & [2.04, 2.60] & 45\% & 4\% & 37\% \\
12 & \texttt{claude-3-haiku} & 2.29 & [1.77, 2.86] & 44\% & 4\% & 52\% \\
13 & \texttt{llama-3.3-70b} & 2.13 & [1.82, 2.43] & 45\% & 1\% & 24\% \\
14 & \texttt{gemini-2.5-pro} & 2.11 & [1.89, 2.35] & 51\% & 1\% & 28\% \\
15 & \texttt{nemotron-3-nano-30b-a3b} & 2.09 & [1.84, 2.32] & 47\% & 1\% & 35\% \\
16 & \texttt{gpt-3.5-turbo} & 2.06 & [1.83, 2.31] & 40\% & 2\% & 33\% \\
17 & \texttt{llama-4-maverick} & 2.05 & [1.74, 2.40] & 44\% & 1\% & 20\% \\
18 & \texttt{sonar} & 2.03 & [1.72, 2.38] & 42\% & 5\% & 25\% \\
19 & \texttt{mythomax-l2-13b} & 2.02 & [1.72, 2.32] & 40\% & 2\% & 25\% \\
20 & \texttt{gpt-5.6-terra} & 1.99 & [1.73, 2.27] & 51\% & 1\% & 20\% \\
21 & \texttt{gemini-3-flash-preview} & 1.98 & [1.69, 2.30] & 46\% & 1\% & 19\% \\
22 & \texttt{qwen-2.5-72b} & 1.98 & [1.73, 2.25] & 40\% & 2\% & 29\% \\
23 & \texttt{claude-sonnet-5.5} & 1.98 & [1.69, 2.29] & 43\% & 0\% & 19\% \\
24 & \texttt{hermes-3-llama-3.1-405b} & 1.95 & [1.68, 2.25] & 42\% & 1\% & 25\% \\
25 & \texttt{gpt-5.6-sol} & 1.94 & [1.70, 2.20] & 42\% & 1\% & 23\% \\
26 & \texttt{command-a} & 1.94 & [1.70, 2.20] & 44\% & 1\% & 24\% \\
27 & \texttt{deepseek-r1} & 1.92 & [1.70, 2.15] & 42\% & 2\% & 35\% \\
28 & \texttt{gpt-4o-mini} & 1.92 & [1.69, 2.16] & 45\% & 1\% & 25\% \\
29 & \texttt{kimi-k2} & 1.92 & [1.68, 2.15] & 39\% & 1\% & 29\% \\
30 & \texttt{nemotron-3.5-lightning} & 1.90 & [1.69, 2.12] & 39\% & 3\% & 31\% \\
31 & \texttt{deepseek-v4-flash} & 1.89 & [1.67, 2.10] & 38\% & 3\% & 32\% \\
32 & \texttt{gpt-oss-20b} & 1.88 & [1.66, 2.11] & 41\% & 1\% & 32\% \\
33 & \texttt{gemma-3-27b} & 1.82 & [1.57, 2.09] & 39\% & 2\% & 19\% \\
34 & \texttt{grok-4.20} & 1.82 & [1.55, 2.12] & 32\% & 2\% & 23\% \\
35 & \texttt{hermes-3-llama-3.1-70b} & 1.81 & [1.54, 2.09] & 38\% & 1\% & 23\% \\
36 & \texttt{deepseek-v4-pro} & 1.81 & [1.58, 2.05] & 36\% & 2\% & 30\% \\
37 & \texttt{claude-fable-5} & 1.76 & [1.50, 2.02] & 41\% & 0\% & 20\% \\
38 & \texttt{granite-4.2-8b} & 1.72 & [1.49, 1.96] & 41\% & 1\% & 26\% \\
39 & \texttt{gpt-4-turbo} & 1.70 & [1.45, 1.97] & 38\% & 0\% & 22\% \\
40 & \texttt{granite-4.2-30b} & 1.69 & [1.44, 1.93] & 43\% & 1\% & 27\% \\
41 & \texttt{claude-opus-5} & 1.67 & [1.45, 1.91] & 40\% & 1\% & 20\% \\
42 & \texttt{gemini-2.5-flash} & 1.63 & [1.41, 1.88] & 37\% & 1\% & 24\% \\
43 & \texttt{gpt-4o} & 1.63 & [1.43, 1.84] & 38\% & 0\% & 27\% \\
44 & \texttt{deepseek-v3-0324} & 1.61 & [1.39, 1.84] & 36\% & 0\% & 26\% \\
45 & \texttt{gpt-6-luna} & 1.60 & [1.39, 1.82] & 37\% & 0\% & 26\% \\
46 & \texttt{gpt-5.6-luna} & 1.60 & [1.41, 1.79] & 37\% & 0\% & 25\% \\
47 & \texttt{qwen3.8-27b} & 1.57 & [1.36, 1.79] & 40\% & 1\% & 26\% \\
48 & \texttt{ling-3.0-flash} & 1.56 & [1.35, 1.77] & 38\% & 0\% & 27\% \\
49 & \texttt{kimi-k2.5} & 1.56 & [1.37, 1.76] & 37\% & 1\% & 27\% \\
50 & \texttt{gpt-5.4-mini} & 1.54 & [1.34, 1.75] & 33\% & 0\% & 27\% \\
51 & \texttt{claude-haiku-4.5} & 1.54 & [1.32, 1.76] & 30\% & 0\% & 23\% \\
52 & \texttt{palmyra-x5} & 1.51 & [1.30, 1.74] & 30\% & 0\% & 20\% \\
53 & \texttt{glm-5.3-flash} & 1.51 & [1.31, 1.72] & 33\% & 1\% & 24\% \\
54 & \texttt{gpt-5} & 1.47 & [1.28, 1.67] & 35\% & 0\% & 22\% \\
55 & \texttt{mimo-v2.6-pro} & 1.47 & [1.31, 1.63] & 34\% & 0\% & 28\% \\
56 & \texttt{gemma-4-31b-it} & 1.46 & [1.26, 1.66] & 39\% & 0\% & 18\% \\
57 & \texttt{muse-glimmer-30b} & 1.46 & [1.26, 1.67] & 38\% & 0\% & 24\% \\
58 & \texttt{qwen3.5-9b} & 1.45 & [1.27, 1.64] & 38\% & 0\% & 24\% \\
59 & \texttt{grok-4.3} & 1.44 & [1.27, 1.63] & 40\% & 0\% & 24\% \\
60 & \texttt{claude-sonnet-4.6} & 1.44 & [1.23, 1.67] & 38\% & 0\% & 17\% \\
61 & \texttt{gpt-4.1} & 1.43 & [1.23, 1.63] & 31\% & 0\% & 21\% \\
62 & \texttt{granite-4.1-8b} & 1.41 & [1.05, 1.80] & 30\% & 1\% & 34\% \\
63 & \texttt{gpt-6-astra} & 1.41 & [1.24, 1.60] & 35\% & 0\% & 20\% \\
64 & \texttt{qwen3.6-35b-a3b} & 1.41 & [1.21, 1.63] & 36\% & 0\% & 25\% \\
65 & \texttt{gpt-5.4} & 1.39 & [1.22, 1.57] & 36\% & 0\% & 21\% \\
66 & \texttt{glm-5.3} & 1.39 & [1.22, 1.57] & 31\% & 0\% & 24\% \\
67 & \texttt{nemotron-3-super-120b-a12b} & 1.38 & [1.21, 1.56] & 32\% & 0\% & 26\% \\
68 & \texttt{ernie-4.5-vl} & 1.38 & [1.15, 1.62] & 34\% & 0\% & 14\% \\
69 & \texttt{gpt-6-sol} & 1.37 & [1.18, 1.56] & 32\% & 0\% & 23\% \\
70 & \texttt{minimax-m3} & 1.37 & [1.18, 1.57] & 28\% & 0\% & 23\% \\
71 & \texttt{seed-2.0-pro} & 1.37 & [1.16, 1.59] & 28\% & 1\% & 20\% \\
72 & \texttt{claude-sonnet-4.5} & 1.35 & [1.15, 1.56] & 37\% & 0\% & 17\% \\
73 & \texttt{step-3.7-flash} & 1.35 & [1.16, 1.56] & 32\% & 1\% & 27\% \\
74 & \texttt{claude-opus-5.5} & 1.35 & [1.15, 1.55] & 31\% & 0\% & 17\% \\
75 & \texttt{kimi-k3} & 1.35 & [1.18, 1.53] & 30\% & 1\% & 22\% \\
76 & \texttt{qwen3.5-122b-a10b} & 1.35 & [1.15, 1.55] & 36\% & 0\% & 22\% \\
77 & \texttt{gemini-3.1-pro} & 1.34 & [1.16, 1.53] & 36\% & 0\% & 21\% \\
78 & \texttt{gemma-4-26b-a4b-it} & 1.33 & [1.15, 1.52] & 36\% & 0\% & 17\% \\
79 & \texttt{claude-opus-4.1} & 1.32 & [1.10, 1.56] & 32\% & 0\% & 18\% \\
80 & \texttt{glm-4.7} & 1.32 & [1.16, 1.51] & 33\% & 0\% & 25\% \\
81 & \texttt{gpt-oss-120b} & 1.32 & [1.15, 1.50] & 30\% & 0\% & 21\% \\
82 & \texttt{hy4-preview} & 1.31 & [1.15, 1.49] & 31\% & 0\% & 25\% \\
83 & \texttt{qwen3.7-plus} & 1.30 & [1.13, 1.48] & 36\% & 0\% & 22\% \\
84 & \texttt{claude-fable-5.1} & 1.30 & [1.13, 1.49] & 30\% & 0\% & 18\% \\
85 & \texttt{gpt-5.5} & 1.28 & [1.12, 1.45] & 30\% & 0\% & 22\% \\
86 & \texttt{qwen3.5-27b} & 1.28 & [1.10, 1.45] & 35\% & 0\% & 21\% \\
87 & \texttt{qwen3-235b} & 1.25 & [1.06, 1.45] & 26\% & 0\% & 19\% \\
88 & \texttt{nemotron-3-ultra-550b-a55b} & 1.25 & [1.08, 1.42] & 28\% & 0\% & 21\% \\
89 & \texttt{gpt-6.1-sol} & 1.24 & [1.09, 1.41] & 30\% & 0\% & 21\% \\
90 & \texttt{gemini-3.1-flash-lite} & 1.24 & [1.07, 1.41] & 30\% & 0\% & 17\% \\
91 & \texttt{gemini-3.5-flash} & 1.24 & [1.08, 1.41] & 33\% & 0\% & 19\% \\
92 & \texttt{claude-opus-4.8} & 1.21 & [1.04, 1.40] & 33\% & 0\% & 17\% \\
93 & \texttt{gemini-3.8-flash} & 1.20 & [1.03, 1.38] & 30\% & 0\% & 20\% \\
94 & \texttt{qwen3.8-2.4t-a95b} & 1.20 & [1.04, 1.36] & 32\% & 0\% & 20\% \\
95 & \texttt{grok-4.7} & 1.20 & [1.04, 1.37] & 27\% & 0\% & 22\% \\
96 & \texttt{grok-4.6} & 1.19 & [1.03, 1.36] & 30\% & 0\% & 21\% \\
97 & \texttt{claude-sonnet-5} & 1.19 & [1.03, 1.35] & 26\% & 0\% & 18\% \\
98 & \texttt{inkling} & 1.19 & [1.02, 1.35] & 31\% & 0\% & 20\% \\
99 & \texttt{inkling-small} & 1.19 & [1.02, 1.37] & 30\% & 0\% & 22\% \\
100 & \texttt{grok-4.5} & 1.17 & [1.01, 1.33] & 30\% & 0\% & 20\% \\
101 & \texttt{muse-spark-1.3} & 1.14 & [0.99, 1.29] & 27\% & 0\% & 19\% \\
102 & \texttt{claude-opus-4.6} & 1.14 & [0.97, 1.30] & 28\% & 0\% & 15\% \\
103 & \texttt{gemini-3.6-flash} & 1.13 & [0.99, 1.28] & 30\% & 0\% & 18\% \\
104 & \texttt{claude-opus-4.7} & 1.12 & [0.96, 1.28] & 32\% & 0\% & 15\% \\
105 & \texttt{claude-opus-4.5} & 1.00 & [0.86, 1.15] & 21\% & 0\% & 15\% \\
\bottomrule

\end{longtable}
}

\section{Prompts}
\label{app:prompts}
The 96 prompts, from the frozen stimulus files: the first 31 from \texttt{spec/stimulus.json} and the rest from \texttt{spec/stimulus\_expanded.json}. Each is followed by \emph{``Reply with one word only.''} and sent as a single-turn conversation with no system prompt, requested temperature 1.0, \texttt{max\_tokens} 1024, eight independent runs per model.

\begin{multicols}{3}
\small
\begin{itemize}\setlength{\itemsep}{0pt}
\item Name a color.
\item Name an animal.
\item Name a fruit.
\item Name a vegetable.
\item Name a city.
\item Name a country.
\item Name a flower.
\item Name a sport.
\item Name a musical instrument.
\item Name a bird.
\item Name a gemstone.
\item Name a tree.
\item Name a beverage.
\item Name an insect.
\item Name an occupation.
\item Name a language.
\item Name a dessert.
\item Name a tool.
\item Name a fish.
\item Name a metal.
\item Name a fabric.
\item Name an herb.
\item Name a dance.
\item Name a hobby.
\item Name a condiment.
\item Name a cheese.
\item Name a dinosaur.
\item Name a mythical creature.
\item Name a board game.
\item Name an emotion.
\item Pick a word.
\item Name a scientist.
\item Name an inventor.
\item Name a painter.
\item Name a composer.
\item Name an author.
\item Name a poet.
\item Name a philosopher.
\item Name a historical leader.
\item Name a hero.
\item Name a villain.
\item Name a landmark.
\item Name a river.
\item Name an island.
\item Name a lake.
\item Name a spice.
\item Name a food.
\item Name a pasta.
\item Name a berry.
\item Name a nut.
\item Name a soup.
\item Name a bread.
\item Name a grain.
\item Name a cocktail.
\item Name a snack.
\item Name a candy.
\item Name a sauce.
\item Name a vehicle.
\item Name a piece of furniture.
\item Name a weapon.
\item Name a hat.
\item Name a shape.
\item Name an article of clothing.
\item Name a material.
\item Name a toy.
\item Name a type of building.
\item Name a type of weather.
\item Name a rock.
\item Name a type of cloud.
\item Name a chemical element.
\item Name a sea creature.
\item Name a dog breed.
\item Name a reptile.
\item Name a mushroom.
\item Name a mammal.
\item Name a plant.
\item Name a virtue.
\item Name a vice.
\item Name a superpower.
\item Name a religion.
\item Name a car brand.
\item Name a holiday.
\item Name a company.
\item Name a brand.
\item Name a sports team.
\item Name a currency.
\item Name a smell.
\item Name a sound.
\item Name a texture.
\item Name a mountain.
\item Name a desert.
\item Name a mineral.
\item Name a constellation.
\item Name a kitchen utensil.
\item Name a musical genre.
\item Name a mythological god.

\end{itemize}
\end{multicols}

\end{document}